\documentclass{article}

\usepackage[numbers]{natbib}
\usepackage[final]{ap_2021}

\usepackage[utf8]{inputenc} % allow utf-8 input
\usepackage[T1]{fontenc}    % use 8-bit T1 fonts
\usepackage{hyperref}       % hyperlinks
\usepackage{url}            % simple URL typesetting
\usepackage{booktabs}       % professional-quality tables
\usepackage{amsfonts}       % blackboard math symbols
\usepackage{nicefrac}       % compact symbols for 1/2, etc.
\usepackage{microtype}      % microtypography

\usepackage{graphicx}
\usepackage{mathtools, nccmath}
\usepackage{tablefootnote}
\usepackage{times}
\usepackage{latexsym}
\usepackage{url}
\usepackage{tabularx}
\usepackage{algorithm}
\usepackage{algpseudocode}

\title{CASPR: A Commonsense Reasoning-based Conversational Socialbot}

\author{
	{\bf Kinjal Basu, Huaduo Wang}, \\{\bf Nancy Dominguez, Xiangci Li, Fang Li, Sarat Chandra Varanasi, Gopal Gupta} \\
	Department of Computer Science\\
	The University of Texas at Dallas\\
	 \{Kinjal.Basu, Huaduo.Wang, Nancy.Dominguez, Xiangci.Li, Fang.Li, \\ Sarat-Chandra.Varanasi, Gopal.Gupta\}@utdallas.edu
}

\begin{document}

\maketitle

\begin{abstract}
We report on the design and development of the CASPR system, a socialbot designed to compete in the Amazon Alexa Socialbot Challenge 4. CASPR's distinguishing characteristic is that it will use automated commonsense reasoning to truly \textit{``understand''} dialogs, allowing it to converse like a human. Three main requirements of a socialbot are that it should be able to ``understand'' users' utterances, possess a strategy for holding a conversation, and be able to learn new knowledge. We developed techniques such as conversational knowledge template (CKT) to approximate commonsense reasoning needed to hold a conversation on specific topics. We present the philosophy behind CASPR's design as well as details of its implementation. We also report on CASPR's performance as well as discuss lessons learned.
\end{abstract}

\section{Introduction}
%test citation \cite{asp_book}

%Pts to mention: 
%1. We are reasoning based, emphasis is on machine learning based approaches
%2. Why ML based approach has limitations

Conversational AI has been an active area of research, starting from a rule-based system, such as ELIZA \cite{eliza} and PARRY \cite{parry}, to the recent open domain, data-driven conversational agents like Amazon's Alexa. Early rule-based bots were based on just syntax analysis, while today's bots are based on advanced machine learning (ML) and deep learning technologies \cite{chatbot-ref}. 

A realistic socialbot should be able to understand and reason like a human. In human to human conversations, we do not always tell every detail, we expect the listener to fill gaps through their commonsense knowledge and commonsense reasoning abilities. This implies that knowledge representation and commonsense reasoning should play an important role in the development of socialbots and chatbots. However, today's systems largely rely on machine learning and deep learning technologies. One of the challenge of modern ML-based chatbots is the lack of ``understanding'' of the conversation. Current chatbots learn patterns in data from large corpora and compute a response without having any semantical grounding of the knowledge internally. 
%That restricts them to provide explanation of a response and shows their limitations.

Our objective in this project is to build a socialbot (called CASPR) that can hold a conversation with an arbitrary person about general topics such as movies, books, music, family, pet, travel, sports, etc., using the Amazon CoBot framework. Our intent is to have CASPR converse in a manner similar to how humans hold conversations, where a human has a very good understanding of the other person's utterances.
An effective socialbot must be able to ``understand'' both explicit and implicit knowledge entailed in a human utterance, be able to plan a response using the knowledge entailed in the sentence augmented with the commonsense knowledge it possesses, as well as be able to learn new knowledge. Our CASPR socialbot is indeed designed with these three aspects in mind.

Our original plan was to use \textit{answer set programming} (ASP) as the underlying technology since it can support commonsense knowledge representation and reasoning, allowing us to automate the human thought process. We have used ASP in the past for developing many intelligent applications where human thought process had to be emulated \cite{chef-zhuo,aaai21,aqua,basu2020square}. Our plan also included using the s(CASP) goal-directed Answer Set Programming system \cite{arias2018constraint} that we have developed, particularly, for automating commonsense reasoning, as it provides significant advantages over standard SAT-solver-based implementations, especially, for developing conversational bots \cite{aaai21}. Due to various limitations, such as issues with latency in making commonsense inferences using the s(CASP) system, need to represent a large amount of commonsense knowledge, we had to approximate the processes used in ASP/s(CASP) and implement them in Python in the CoBot framework. Due to these limitations and given the competition's tight schedule, CASPR could not quite reach its goal of completely ``understanding'' the knowledge entailed in a user's utterance. Instead of completely understanding the utterance, CASPR organizes the dialogs in such a way, where possible, that it can anticipate what a user is going to say. Of course, our research into realizing chatbots and socialbots that are based on commonsense reasoning and truly ``understanding'' users' utterances continues \cite{aaai21,discasp}.

%Rest of the paper is organized as follows. %GG HERE

\section{Background}

We next give a brief introduction to background concepts and technologies.

\subsection{Commonsense Reasoning}

As mentioned earlier, a realistic socialbot should be able to understand and reason like a human. In human to human conversations, we do not always tell every detail, we expect the listener to fill gaps through their commonsense knowledge and commonsense reasoning. Thus, to obtain a conversational bot, we need to automate commonsense reasoning, i.e., automate the human thought process. The human thought process is flexible and \textit{non-monotonic} in nature, which means \textit{ ``what we believe today may become false in the future with new knowledge''}. We can model commonsense reasoning with (i) default rules, (ii) exceptions to defaults, (iii) preferences over multiple defaults \cite{gelfond1988stable}, and (iv) modeling \textit{multiple worlds} \cite{gelfond2014knowledge,baral,gupta-csr}. 

Much of human knowledge consists of default rules, for example, the rule: \textit{Normally, birds fly}. However, there are exceptions to defaults, for example, \textit{penguins are exceptional birds that do not fly}. Reasoning with default rules is non-monotonic, as a conclusion drawn using a default rule may have to be withdrawn if more knowledge becomes available and the exceptional case applies. For example, if we are told that Tweety is a bird, we will conclude it flies. Later, knowing that Tweety is a penguin will cause us to withdraw our earlier conclusion.

Humans often make inferences in the absence of complete information. Such an inference may be revised later as more information becomes available. This human-style reasoning is elegantly captured by default rules and exceptions. Preferences are needed when there are multiple default rules, in which case additional information gleaned from the context is used to resolve which rule is applicable. One could argue that expert knowledge amounts to learning defaults, exceptions and preferences in the field that a person is an expert in.

Also, humans can naturally deal with \textit{multiple worlds}. These worlds may be consistent with each other in some parts, but inconsistent in other parts. For example, animals don't talk like humans in the real world, however, in the cartoon world, animals do talk like humans. So,  a fish called Nemo, may be able to swim in both the real world and the cartoon world, but can only talk in the cartoon world. Humans have no trouble separating cartoon world from real world and switching between the two as the situation demands. Default reasoning augmented with the ability to operate in multiple worlds, allows one to closely represent the human thought process \cite{gupta-csr}. Default rules with exceptions and preferences and multiple worlds can be elegantly realized with answer set programming \cite{gelfond2014knowledge,baral} and the s(CASP) system \cite{arias2018constraint}. 

\subsection{Answer Set Programming}

Answer Set Programming (ASP) is a declarative paradigm that extends logic programming with negation-as-failure. ASP is a highly expressive paradigm that can elegantly express complex reasoning methods, including those used by humans, such as default reasoning, deductive and abductive reasoning, counterfactual reasoning, constraint satisfaction, etc. ~\cite{baral,gelfond2014knowledge,gupta-csr}.

ASP supports better semantics for negation ({\it negation as failure}) than does standard logic programming and Prolog. An ASP program consists of rules that look like Prolog rules. The semantics of an ASP program {$\Pi$} is given in terms of the answer sets of the program \texttt{ground($\Pi$)}, where \texttt{ground($\Pi$)} is the program obtained from the substitution of elements of the \textit{Herbrand universe} for variables in $\Pi$~\cite{baral}.
The rules in an ASP program are of the form:
\begin{center}
  \texttt{p :- q$_1$, ..., q$_m$, not r$_1$, ..., not r$_n$.}
\end{center}
\noindent where $m \geq 0$ and $n \geq 0$. Each of \texttt{p} and \texttt{q$_i$} ($\forall i \leq m$) is a literal, and each \texttt{not r$_j$} ($\forall j \leq n$) is a \textit{naf-literal} (\texttt{not} is a logical connective called \textit{negation-as-failure} or \textit{default negation}). The literal \texttt{not r$_j$} is true if proof of {\tt r$_j$} \textit{fails}. Negation as failure allows us to take actions that are predicated on failure of a proof. Thus, the rule {\tt r :- not s.} states that {\tt r} can be inferred if we fail to prove {\tt s}. Note that in the rule above, {\tt p} is optional. Such a headless rule is called a constraint, which states that conjunction of {\tt q$_i$}'s and \texttt{not r$_j$}'s should yield \textit{false}. Thus, the constraint {\tt :- u, v.} states that {\tt u} and {\tt v} cannot be both true simultaneously in any model of the program (a model is called an answer set).

The declarative semantics of an Answer Set Program $\Pi$  is given via the Gelfond-Lifschitz transform~\cite{baral,gelfond2014knowledge} in terms of the answer sets of the program \texttt{ground($\Pi$)}. ASP also supports classical negation. A classically negated predicate (denoted {\tt -p}) means that {\tt p} is definitely false. Its definition is no different from a positive predicate, in that explicit rules have to be given to establish {\tt -p}. More details on ASP can be found elsewhere~\cite{baral,gelfond2014knowledge}. 

The goal in ASP is to compute an {\it answer set} given an answer set program, i.e., compute the set that contains all propositions that if set to true will serve as a model of the program (those propositions that are not in the set are assumed to be false).  Intuitively, the rule above says that {\tt p} is in the answer set if {\tt q$_1$, ..., q$_m$} are in the answer set and {\tt r$_1$, ..., r$_n$} are not in the answer set. 
ASP can be thought of as Prolog extended with a sound semantics of negation-as-failure that is based on the stable model semantics~\cite{gelfond1988stable}.

\subsection{The s(CASP) System}

Considerable research has been done on answer set programming since the inception of the stable model semantics that underlies it~\cite{gelfond2014knowledge}. A major problem with ASP implementations is that programs have to be grounded and SAT-solver-based implementations such as CLASP~\cite{clingo} used to execute the propositionalized program to find the answer sets. There are multiple problems with this SAT-based implementation approach, which includes exponential blowup in program size, having to compute the entire model, and not being able to  produce a justification for a conclusion~\cite{arcade}. Goal-directed implementations of ASP~\cite{sasp,arias2018constraint}, called s(ASP) and s(CASP), work directly on predicate ASP programs (i.e., no grounding is needed) and are query-driven (similar to Prolog). The s(ASP) and s(CASP) systems only explore the parts of the knowledge-base that are needed to answer the query, and they provide a proof tree that serves as justification for the query. The s(ASP) and s(CASP) systems support predicates with arbitrary terms as arguments as well as constructive negation \cite{sasp,arias2018constraint}.
 
%In the work reported here, we will mainly use the s(CASP) system that additionally supports constraint solving over reals, which is important for reasoning faithfully about (continuous) time. The s(CASP) system is the key technology for representing and analyzing CLEAR requirements modeled with the event calculus. The s(CASP) system also directly supports abductive reasoning and can provide justification for a query.

\subsection{Deep Learning}
Deep learning \cite{lecun2015deep} is a type of ML method that focuses on artificial neural networks. In the past decade, it is widely applied in natural language processing (NLP), which takes computational approaches to understand, and generate human languages. Since building dialogue systems is all about understanding, processing and generating conversational texts, deep learning naturally has been applied to developing socialbots. A socialbot is divided into several modules that perform standard NLP tasks, such automatic speech recognition (ASR), intent classification, named entity recognition (NER) and response generation. There has been extensive research on each of the task and Amazon's CoBot framework has provided out-of-box implementations for these task modules.

Since the advent of BERT model \cite{devlin2018bert} which archived state-of-the-art performances in a number of NLP tasks such as NER and question-answering, the NLP community has gradually switched to language models (LMs) based on Transformers \cite{vaswani2017attention}. A Transformer model consists of identical transformer blocks, which contain self-attention mechanisms to compute the activation of each input token contextualized by their neighbors' tokens \cite{vaswani2017attention}. Due to the easily parallelizable nature of the attention mechanism, Transformer-based LMs are able to expand to millions \cite{devlin2018bert, liu2019roberta} or even billions \cite{raffel2019exploring} of parameters, so that a newly released model may top the NLP benchmark leaderboards every a few months. As a result, a typical usage of these LMs is to take a top-performing LM and fine-tune it for various tasks.

Despite the great success of the deep learning models such as Transformer-based LMs on standard NLP tasks such as ASR and NER, applying large LMs on dialogue systems faces some challenges. One of the problems is that deep learning models are used as black box with few efficient technique to control the output. As a result, the deep learning lacks consistency and the outputs are less predictable than rule-based systems. Another issue is that deep learning approach requires large datasets, long training and inference time and high computational budget. All these issues discourage us entirely relying on deep learning approaches.

\section{CASPR Design Philosophy}

Our philosophy is to design a socialbot that emulates, as much as possible, the way humans conduct social conversations. Humans employ both learned-pattern matching (e.g., recognizing user sentiments) and commonsense reasoning (e.g., if a user starts talking about having seen the Eiffel Tower, we infer that they must have traveled to France in the past) during a conversation. Thus, ideally, a socialbot should make use of both machine learning as well as commonsense reasoning technologies. Our goal is to use the appropriate technology for a task, i.e., use machine learning and commonsense reasoning for respective tasks that they are good at. Machine learning is good for tasks such as parsing, topic modeling, and sentiment detection while commonsense reasoning is good for tasks such as generating a response to an utterance. In a nutshell, we should use machine learning for modeling \textit{System 1} thinking and commonsense reasoning for modeling \textit{System 2} thinking \cite{kahneman,gupta-csr}. We strongly believe that intelligent systems that emulate human ability should follow this approach, especially, if we desire true understanding and explainability.

A human must possess at least the following three skills in order to hold an engaging social conversation: (i) understand and ``process'' what a user is saying, (ii) plan a conversation strategy, generate responses and pivot through various topics, and (iii) be able to acquire new knowledge for future conversation, e.g., by reading up on a topic. A successful socialbot must have these abilities as well.
Thus, a socialbot must have the ability to understand what a user is saying. This entails syntactically parsing the utterance including performing NER and understanding its semantic content. Understanding semantic content of a user utterance is the toughest challenge. One could use a ML-based approach or a knowledge-representation-based approach. Our goal was to develop an approach based on knowledge representation and commonsense reasoning, as discussed earlier. A socialbot must also be able to understand a user's intent, his/her sentiments, as well as topics the user wants to talk about. These abilities are best realized through the use of machine learning technologies.
a socialbot must also possess conversation planning skills: it (i) must understand how a conversation is organized, i.e., it must have a plan for holding a conversation, (ii) be able to generate a response, and (iii) be able to pivot topics as the conversation evolves. At a very high level, a social conversation can be thought of as discussing through a standard set of topics such as movies, books, music, sports, family, pets, etc., one after the other in some random order. Each topic should last for at least a few dialogs. Most people have an ingrained strategy in their mind for holding a social conversation when they meet a stranger at a party, for example. A socialbot must have a similar strategy at its core. 
Finally, just like a human, a socialbot must possess the ability to acquire new knowledge, i.e., gain knowledge of topics, for example, that may have been brought up by a user and that the socialbot knew nothing about. We next give details of how we realized each of these three abilities.

%our philosophy: emulate a human

\subsection{Conversation Planning}

CASPR's conversation planning is centered around a loop in which it moves from topic to topic, and within a topic, it moves from one attribute of that topic to another. Thus, CASPR has an \textit{outer conversation loop} to hold the conversation at the topmost level and an inner loop in which it moves from attribute to attribute of a topic. The logic of these loops is slightly involved, as a user may return to a topic or an attribute at any time, and CASPR must remember where the user left off in that topic or attribute. For the inner loops, CASPR uses a template, called conversational knowledge template (CKT), that can be used to automatically generate code that loops over the attributes of a topic, or loops through various dialogs (mini-CKT) that need to be spoken by CASPR for a given topic.

%Understand how a conversation is organized (conversation Loop and Conversation Knowledge Template (CKT))

\medskip 
\noindent{\bf Outer Conversation Loop:} We designed an \textit{outer conversation loop} that loops through a list of various topics.  These topics are traversed in random order. In a casual social interaction with a stranger, we tend to use the following plan to hold a conversation. We cycle through various topics via dialogs, until we discover a topic of mutual interest. We spend a few dialog cycles on this topic of mutual interest, then we move on to the next topic, then to another topic, and on and on in an endless cycle. At any moment, the user can ask a random question, make an unrelated comment, or stop. CASPR's outer conversation loop is designed following this plan. This is illustrated in Figure \ref{fig:outer-loop} below. Conversation topics currently include movies, books, music, sports, family, pets, food, travel, school, technology, etc. New topics can be easily added, as described later.  The outer conversation loop can be thought of as an automaton that transitions between various states, where each state corresponds to a topic.  Note that in Figure \ref{fig:outer-loop}, each node is connected to every other node (not all the links are shown so as not to clutter the figure), as we can transition from any topic to any other topic. The orange dots represent a user utterance that is irrelevant to any of the topics and thus has to be handled by other response generators. 

\noindent
\setlength{\belowcaptionskip}{-10pt}
\begin{figure}[h]
    \centering
    \includegraphics[width = \textwidth]{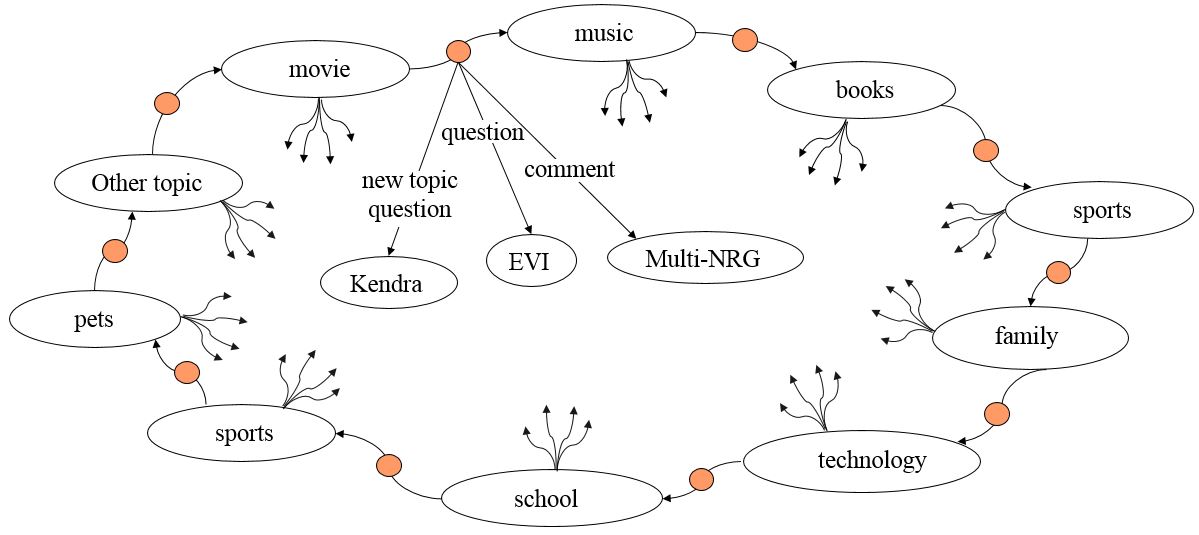}
    \caption{Structure of the Outer Conversation Loop}
    \label{fig:outer-loop}
    \end{figure}

\begin{figure}[ht]
    \centering
    \includegraphics[width = \textwidth]{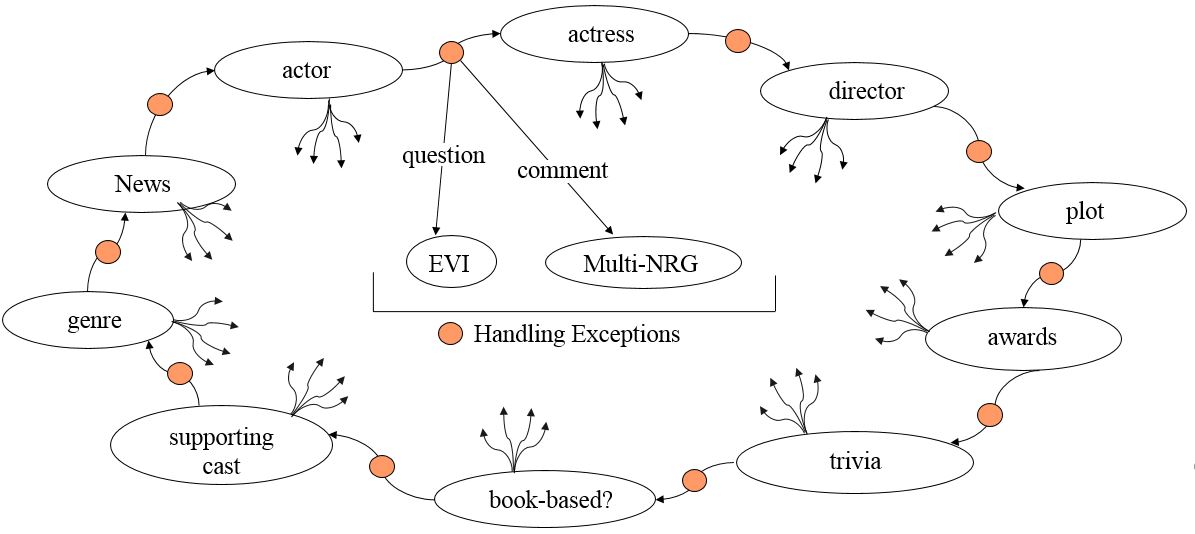}
    \caption{Conversational Knowledge Template with Exceptions}
    \label{fig:movie-ckt}
    \end{figure}

\medskip \noindent{\bf Conversational Knowledge Template with Exceptions:} For each topic, we need to have knowledge about that topic. For movies, books, and music, extensive databases are available that contain pertinent knowledge: IMDB for movies, Goodreads for books, and Spotify for music. For such topics where structured knowledge is available, we designed a system that when given information about a specific movie, book, or album, will find all the relevant additional information from the corresponding database and use that information for holding a conversation about that topic. For example, if we know that a user's favorite movie is Titanic, then all the information about Titanic's actors, director, awards, movie-plot, etc., will be retrieved from IMDB. To facilitate holding a conversation about a given topic, we designed \textit{conversational knowledge templates with exceptions} (CKT). A CKT allows us to hold conversations about a specific topic. The CKTs for movies, music and books are designed using the corresponding databases. The schemata for a movie CKT is shown in Figure \ref{fig:movie-ckt}. Each node represents conversation around the sub-topic that appears as the node label (default behavior). If the user utterance is not related to that sub-topic, the exception handling mechanism (represented by orange dots) is invoked and an appropriate response generated depending on if the user asked a question, or spoke an utterance not related to the sub-topic. Note that our pivoting strategy (see later) will take us from one node (sub-topic) to another random node. Because we can transition from any node in the movie-CKT to any other, the movie-CKT is really a complete graph with every node connected to every other node.

\noindent For all other categories, where an extensive database is not available, we hard-wire the conversation. We call this hard-wired conversation, a mini-CKT. Mini-CKTs can be designed very rapidly. Mini-CKTs provide us a mean for rapidly adding knowledge for specialized topics. We show an example mini-CKT for the topic of ``travel to Italy'' below. Note that there are a total of five dialogs in the mini-CKT, and for each dialog, there are many ways for CASPR to say that dialog. CASPR will randomly choose one of them. Using a different rendering of a dialog each time it is invoked gives CASPR a more human-like feeling as humans do not utter the same words when they repeat a sentence (to conserve space, we do not show all the variants of CASPR's response in the travel mini-CKT below).

\setlength{\belowcaptionskip}{-10pt}
    
{\footnotesize 
\begin{verbatim} 
Dialog 1:
    1. Do you like to travel?
    2. If I could travel, I would explore the world. What about you? 
               Do you like traveling?
    3. .....
Dialog 2:
    1. Good. My favorite place is Italy. Which country is your favorite 
         for travel?
    2. Great. I would definitely visit Italy. Wonderful sights to visit. 
         Do you agree?
    3. .....
Dialog 3:
    1. Italy has wonderful food too. I would eat the gelato icecream 
        every day. Wouldn't you?
    2. The cuisine in Italy is also great. Mouth-watering desserts such as 
        Tiramisu. Don't you agree?
    3. .....
Dialog 4:
    1. Aside from visiting the wonderful cities, I'd like to
        explore the countryside where wine is made. Wouldn't you?
    2. Italy has so much history, like the Coliseum is 2000 years
        old. I would like to see it. Would you want to visit the Coliseum?
    3. .....
Dialog 5:
    1. France is also one of my favorite places, especially Paris with its 
        Eiffel tower. Which country would you want to go to?
    2. I would also like to visit France, with all its museums and palaces. 
        What's your favorite country to visit?
    3. .....
\end{verbatim} 
}

Note that in the last dialog, we asked about the user's favorite country to visit. If we can pick up that country from the user's response through syntactic analysis, then we can invoke the mini-CKT for travel to that country, if we have one. If we don't have a mini-CKT for that country, the outer loop will transition to another conversation topic. Note also that the code that implements a min-CKT can be automatically generated from a high-level specification of the dialogs (discussed later). 

\medskip\noindent {\bf Pivoting Conversation Topics:}
%
%Pivot conversation topics using subject matter being discussed
In actual conversations between people, we pivot from one sub-topic to another sub-topic within the conversation's current topic. For example, while discussing the movie Titanic, we may talk about its various attributes---actors, actresses, director, genre, plot, awards, trivia, etc.---then \textit{pivot to} another popular movie in which one of the actors has also acted or which the director has also directed. For example, while talking about Titanic's director James Cameron, we may pivot to another one of his well-known movies such as Avatar. Our conversation will then drift towards this new movie. We may talk for a few minutes about Avatar, and then pivot to another movie related to Avatar via its genre this time (e.g., talk about another popular science fiction movie such as Wall-E). A socialbot must have a topic pivoting strategy; CASPR's pivoting algorithm is based on using the attributes of a topic. 

%advantage of declarative knowledge

\medskip\noindent{\bf Response Generation:} 
%
%Generate a response to a user utterance (Multiple response generators select the best response)
Ideally, CASPR should ``understand'' a user's utterance (i.e., map the semantics of the spoken sentence to \textit{knowledge}) just as a human does. CASPR should then use that knowledge to find related information or draw new inferences and use them for further advancing the conversation. CASPR's original goal indeed was to represent this knowledge in ASP and then draw new inferences \cite{aaai21,aqua,discasp}. However, due to the diversity of topics involved and the amount of effort needed to model commonsense knowledge, we had to weaken this goal. Instead of ``understanding a user's response, we designed the CKTs and miniCKTs in such a way that we could anticipate what the user would say. Essentially, this meant that CASPR's response almost always ended in a question that the user had to answer. Thus, if the user went along with these questions, then a reasonably smooth conversation resulted. As long as a user's responses were judged to be within the topic being pursued or neutral (e.g., phatic), then CASPR will stay within that topic. If a user asks a question, then the EVI question answering module that is part of the CoBot framework is invoked and the question answered. For any other response from the user, i.e., off-topic and non-question, the Multi-NRG module (see later) is invoked to obtain a response.

\subsection{Understanding User Utterances} 

As discussed earlier, an effective socialbot should be able to ``understand'' user utterances. This entails syntactic parsing as well as semantic mapping to a knowledge representation language such as ASP. The ASP representation can then be used for making further inferences along with additional commonsense knowledge \cite{aaai21,aqua}. For example, if the user says ``Tom Cruise is my favorite actor,'' then from this knowledge, plus the commonsense knowledge that people who like an actor also like their movies, we should be able to infer that the user also likes Tom Cruise movie such as Top Gun. A socialbot that operates like a human must have this capability, however, there are many challenges in achieving this goal. We are assiduously working towards that goal \cite{discasp}, however, at the moment, we have to approximate this knowledge as discussed in the previous section. 
We craft the conversations carefully (through CKTs and mini-CKts) so that we can anticipate what the user is going to say. At present CASPR uses tools such as EVI \cite{EVI} and PD-NRG \cite{NRG} for response generation, so it does rely on machine learning-based tools. These ML-based tools do not really understand the content of the user utterance, rather language models based on machine learning technologies are used for generating responses.

We also use keyword matching to guess what a user is saying. For example, if we find the word ``dog'' in a user utterance, we guess that the user may be interested in talking about dogs and possibly pets. So CASPR will invoke the mini-CKT for a pet.

We also need to understand a user's sentiment and the broad topic of her/his utterance. For this, we use machine learning technologies and language models. Machine learning technologies are ideal for understanding user sentiment and inferring the topic a user wants to talk about, as humans use their intuitive knowledge and their pattern matching abilities for this purpose. Thus, following our philosophy, ML-based tools will be an ideal fit.

%Understand syntactic content of user utterances (Parsers; NER)
    
%Understand the semantic content of user utterances (Dialog management with CKTs (we know what the user is going to say); Recognize specific utterances via regex matching (e.g., we know user is talking about her pet dog); Ideally should use KR and commonsense reasoning)
    
%Understand user intent, sentiment \& topics user is talking about (Machine learning-based tools (topic modeling, user intent, etc.))
        
\subsection{Ability to Learn New Knowledge}

A successful socialbot should be able to learn/ingest new knowledge. In CASPR, this can be accomplished in two ways: 
\begin{itemize}
    \item 
When a new topic is encountered, a new mini-CKT related to that topic can be added. This is however a tedious task because the number of topics can be quite large since we can potentially add a mini-CKT about every narrow topic. The burden of adding a new mini-CKT is reduced by automatically generating the code that implements it, given its high-level specification.
\item Use the Amazon Kendra service \cite{kendra} to add this knowledge. This technique is more scalable as all that is needed is a textual description of the topic and the Kendra service can answer questions against this text after processing this text. The topic modeling module has to be enhanced to recognize the new topic added. 
\end{itemize} 
%New knowledge ingestion (Use Amazon Kendra as well as add CKTs for various topics)

\section{System Architecture}

Our CASPR open-domain chatbot is built atop Amazon's CoBot framework \cite{cobot} that provides a seamless integration of AWS with Alexa Skills Kit (ASK). CoBot is event-driven and it works as a AWS Lambda function in the backend of Alexa Skill (running on ASK). The framework not only helps a developer to quickly build an initial chatbot by using the provided tools, libraries, and base models, but also handles the load balancing and scaling of the components.   

\begin{figure}[h]
    \centering
    \includegraphics[width = \textwidth]{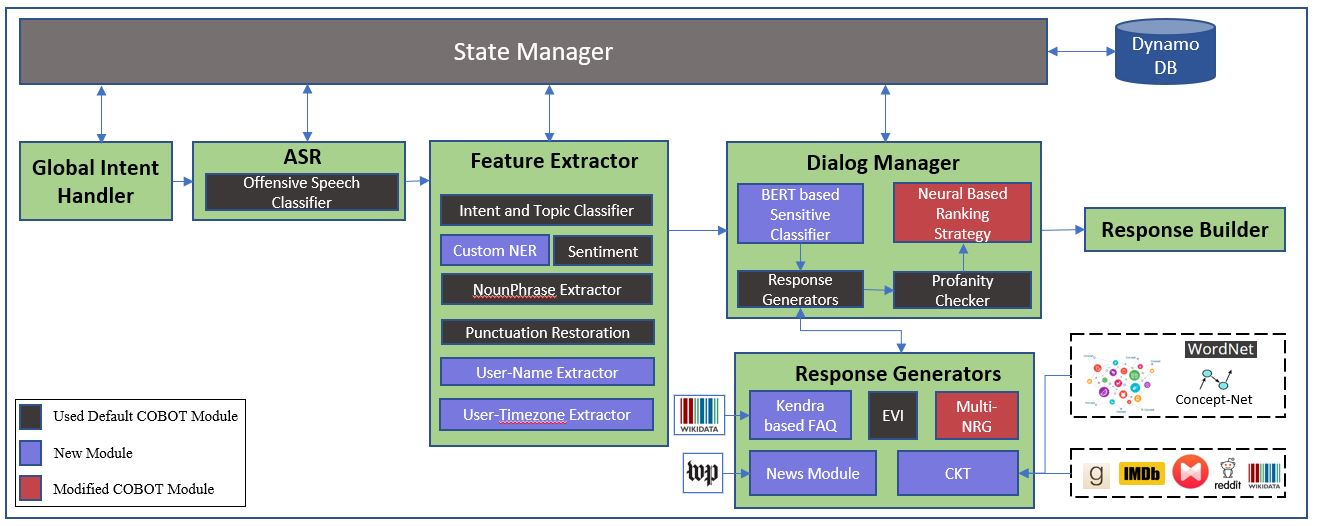}
    \caption{ CASPR Architecture }
    \label{fig:arch}
    \end{figure}

In this section, we explain CASPR's architecture in detail along with the key components. CASPR is made up of seven unique task-based modules that are responsible for creating the seamless end-to-end conversation pipeline for a open-domain chatbot. The unique tasks for each modules are as follows: (i) identification of the dialog intent, (ii) speech to  English text conversion, (iii) NLP feature extraction from the text, (iv) controlling the conversation using the extracted features as well as handling the conversation's sensitivity, (v) crafting the responses based on the topic leveraging external knowledge, (vi) converting the English text response back to speech, and (vii) storing internal processing data in addition to conversation-specific information. Figure \ref{fig:arch} shows CASPR’s architecture. The seven modules are labeled, respectively, global intent handler, automatic speech recognizer (ASR), feature extractor, dialog manager, response generator, response builder, and state manager. For the global intent handler, ASR, and the response builder modules, we  rely heavily on the out-of-the-box tools that come with the CoBot framework and ASK. Our main contribution lies in the feature extraction, dialog management, and the modified/new response generators.

\noindent
\setlength{\belowcaptionskip}{-10pt}

\subsection{Global Intent Handler}
The main job of the global intent handler is to decide whether to end the conversation or continue based on a user utterance. Also, it is responsible for providing the welcome phrase for the bot when a conversation starts. CASPR uses the CoBot's out-of-the-box global intent handler that processes a user's utterance and returns the welcome phrase (if any) and the end-session-flag (true/false).

\subsection{Automatic Speech Recognizer (ASR)}
The built-in Alexa's ASR service has been used in CASPR to translate the user utterances to English text. From our experience with the ASR service, and by manually checking the real public conversations, we found that it really performs well when the utterances are short. The error starts 
%populating 
increasing
when the utterance length increases. Also, errors appeared in handling joined words such as `butter' + `milk' = `buttermilk', where all three words have different meanings. These types of joined words do not occur in a conversation frequently, so this issue arises rarely. From the current conversation experience, we have found that the average length of user utterances is normally small, so ASR errors are not greatly noticeable. However, ASR errors increase when the conversation becomes complex. We will pursue this issue of noisy transcription and error correction in ASR as part of our future work. Our plan is to manually annotate the mislabeled ASR cases to create a dataset that we can use along with the topical-chat dataset \cite{topical_chat} to train a supervised neural model that will post-process the ASR output and correct it.

\subsection{Feature Extractor}
This module is solely responsible for extracting the features from user utterances and that are used later in understanding the utterances.  For the sentiment analysis, noun-phrase extraction, and punctuation restoration, we have used the CoBot framework's out-of-the-box tools. We did experiments with all different tools from the framework for the sentiment analysis and found that `Valence Aware Dictionary and sEntiment Reasoner' (VADER) performs really well.  

\subsubsection{Intent and Topic Classification}
Identification of user intent and the topic a user wants to talk about is very useful for generating an engaging and on-the-topic response. We have experimented with two different intent-topic classifiers. First, we used the ASK-based default classifier where the expected topic classes and the intent classes are defined inside our ASK skill. This classifier uses a rule-based algorithm and automatically classifies the intent and the topic. However, this approach has the drawback of interpreting each user input separately, which generates a lot of false positive classification. In other words, suppose a user has many rounds of conversation on the same topic but due to different phrases the user employs for each utterance, the topic classifier assigns a different topic  to each utterance. This results in the generation of wrong bot responses as the dialog manager are misguided. Next, we tried the CoBot-provided The dialog-Act intent-topic classification that takes the discourse of the conversation as input, which performed much better. This classifier is a trained neural model that predicts the classes from only a given set of 11 topics and 11 intents. Classification using this method is higher level (in the sense of being more abstract) in comparison with the JSON-based classifier, however, it gave rise to our \textit{outer conversation loop} idea.

\subsubsection{Knowledge-driven Named Entity Recognition}

Although the named entity recognition (NER) module provided by Amazon generally performs well on recognizing common named entities, it sometimes fails in recognizing specific entities such as particular movie names when we attempt to understand user utterances. Therefore, we developed additional mechanisms to increase the recall of NER. Since, initially, we limited ourselves to a limited number of topics, such as movies, music, books, we built a NER module for each such topic. Due to our limited budget for GPUs for computing neural network-based NER models, we chose a lightweight information retrieval approach from pre-collected entity corpus augmented with some speed-up techniques. %GG21: perhaps citation needed?

\noindent\textbf{Overall Procedure:} We first collect entity names from datasets that contain entity information from publicly available websites such as IMDb or Kaggle. Then we perform string matching of the collected entity names with user utterances to extract candidate named entities. We leverage some techniques described below to improve the matching performance. We rank the candidate entities by each of their corresponding attribute that reflects the likelihood of them being mentioned by users. For example, we use the number of votes attributed to rank the candidate movie names. 

\noindent\textbf{Entity collection: }
We choose several most frequently mentioned conversation topics that require NER and implement one NER module for each of these entity types. As Table \ref{tab:entity} shows, we extract movie, movie actors, books, book authors, music, music artists, video games as well as sports, including American football players and teams, soccer players and clubs, NBA players, and teams, baseball players and teams, and tennis players from publicly available datasets.

\noindent\textbf{Building Search Index: }
We prepare a search index implemented by Python Dictionary that supports fuzzy search. First, converting each entity to spoken English by removing punctuations, normalizing numerals (Arabic and Roman) to English, and reversing abbreviations (e.g. Mr. to mister). Then we build the entity retrieval index by tokenizing each N-gram entities to k-gram tokens ($k \in {1,2,...,N}$) and map each of the tokens to the original entity name to allow partial matching. Additionally, we also support keys with singletons to retrieve plurals and keys with missing stop-words. To reduces the false positive results, we only keep those k-gram keys with $k/N \geq 0.5$.

\noindent\textbf{Named Entity Retrieval: }
During each run of named entity retrieval, user utterances are normalized and tokenized into a set of k-gram tokens ($k \in {1,2,...,N}$). We ignore certain words with Part-of-Speech tags such as \{WP, PRP, VB\}. We use these k-gram tokens as keys to retrieve candidate entities from the prepared search index. For each of the normalized entities, we compute its longest common sub-sequence with the normalized user utterance. We use the output sub-sequence to compute the matching score $S$, which is the proportion of the matching string length to the candidate entity length. Finally, we rank the candidate entities by considering $S$, length of the candidate entity $L$, and ranking attribute $R$ shown in Table \ref{tab:entity} with the heuristic functions $h_{IMDB}$ for movie and video games, and $h$ for other applicable topics:

\begin{equation*}
\begin{aligned}
h &= S \times \sqrt{L} \times \sqrt{R} \\
h_{IMDB} &= S \times \sqrt{L} \times \log{R} 
\end{aligned}
\end{equation*}

\begin{table}[t]
\small
\begin{center}
    \begin{tabular}{|c|c|c|c|}
    \hline
    \textbf{Topic} & \textbf{Entity Type} & \textbf{Ranking attribute} & \textbf{Source} \\ \hline
    Movie & Movie & Number of votes & IMDb \tablefootnote{\url{https://www.imdb.com/interfaces/}} \\ \hline
    Movie & Movie actor name & Sum of votes of & \\ && movies starred in & IMDb \\ \hline
    Book & Book & Ratings count & Goodreads-books \tablefootnote{\url{https://www.kaggle.com/jealousleopard/goodreadsbooks}} \\ \hline
    Book & Book author & Sum of ratings & \\ && count of works & Goodreads-books \\ \hline
    Music & Music & Popularity & Spotify dataset \tablefootnote{\url{https://www.kaggle.com/yamaerenay/spotify-dataset-19212020-160k-tracks}} \\ \hline
    Music & Music artists & Sum of popularity & \\ && of works & Spotify dataset \\ \hline
    Video games & Video game name & Number of votes & IMDb \\ \hline
    American football & American football player & Experience & NFL statistics \tablefootnote{\url{https://www.kaggle.com/kendallgillies/nflstatistics}} \\ \hline
    American football & American football team & N/A & NFL statistics \\ \hline
    Soccer & Soccer player & Rating & Soccer players statistics \tablefootnote{\url{https://www.kaggle.com/antoinekrajnc/soccer-players-statistics}} \\ \hline
    Soccer & Soccer club & Sum of player ratings & Soccer players statistics \\ \hline
    Basketball & NBA player & Minutes of play & Basketball Players \\ &&& Stats per Season \tablefootnote{\url{https://www.kaggle.com/jacobbaruch/basketball-players-stats-per-season-49-leagues}} \\ \hline
    Basketball & NBA team & N/A & NBA official site \tablefootnote{\url{https://www.nba.com/teams}} \\ \hline
    Baseball & Baseball player & Points won / points max & Baseball databank \tablefootnote{\url{https://www.kaggle.com/open-source-sports/baseball-databank}} \\ \hline
    Baseball & MLB team & N/A & List of MLB teams \tablefootnote{\url{https://www.ducksters.com/sports/list_of_mlb_teams.php}} \\ \hline
    Tennis & Tennis player & Winner rank & WTA/ATP Tennis Data \tablefootnote{\url{https://www.kaggle.com/taylorbrownlow/atpwta-tennis-data?select=KaggleMatches.csv}} \\ \hline
    \end{tabular}
    \caption{Entity collection} \label{tab:entity}
\end{center}
\end{table}

To our best of our knowledge, there is only one movie-related NER dataset that is applicable to our NER modules \footnote{\url{https://groups.csail.mit.edu/sls/downloads/movie/}}. We evaluated our movie title NER and movie actor NER modules on the test and obtained the recall of 0.633 and 0.607, respectively. More annotated NER corpora are needed for improving the results.

\subsubsection{User's Name and Timezone Extraction}
A socialbot is like an AI friend and we all want our friends to remember us. To memorize someone we start from his/her name and then we gradually collect their personal preferences as we talk more and more with them. Also, we always like our friends to remember our choices. So, to provide a personalized experience for returning users, CASPR asks the users their name and stores it in the Dynamo-DB with the unique user's device ID. At anytime later, if the user initiates another conversation with CASPR, it greets the user with his/her name. This definitely gives a good conversation experience to the user, as reflected in the results. Our future plan is to store the user's topic preferences and all the likes/dislikes. Using this information, we can train a supervised model to predict the topics that the user would want to talk about. 

Similarly, CASPR is also capable of greeting a user based on the current time in his/her timezone (morning, afternoon, evening). Given a device ID of a user, CASPR can fetch the current timezone of the device leveraging the Alexa API. Then it uses a python library of timezone called Pytz to get the current time. We were also planning to have a conversation based on the current weather. However, due to the privacy policy, we could not access a user's current location.   

\subsection{Dialog Manager}
This is the brain of the system, which controls the conversation. It starts with identifying sensitive user responses, if it gets any then it returns canned responses to mitigate the tension and to keep the conversation going. Otherwise, it invokes the response generators based on the intent and the topic. Also, it is responsible to  filter the sensitive candidate responses and rank them using a rule or neural based ranking algorithm. Finally, it passes the bot response to the response builder. Following are the details of these components. 

\subsubsection{Sensitive Utterance Filtering}
According to the Alexa Prize regulations, the bots are obligated to filter out not only offensive utterances but also potentially sensitive utterances related to alcohol, disability, finance, law, emergency, medicine, politics, psychology, religion, sex, society, and violence. We came up with a comprehensive black list of sensitive phrases in the categories above. However, it is usually overkill to treat every single utterance that includes these words as a sensitive utterance. For example, users become frustrated when the bot refuses to continue the conversation because ``James Bond'' was mentioned and bond is a financial term. Therefore, our work on sensitive utterance filtering mainly focused on relaxing the filtering condition to mitigate the problem of a high false-positive filtering rate.

\noindent\textbf{White-listing: }
The most intuitive solution is white-listing those sensitive phrases that are not semantically sensitive in utterances. We manually curated these phrases from the recorded conversation history. Some example phrases are ``share'', ``oh my god'' and ``Chinese food''. We also white-list COVID-19 related phrases and process COVID-19 related utterances in our COVID module.

\noindent\textbf{Factual questions: }
Inspired by \cite{paranjape2020neural}, we replicate their solution by training a RoBERTa-base model \cite{liu2019roberta} on a simplified version of a dialog act dataset as a binary classification problem to predict whether a question utterance is a factual question or an open-ended question. We regard all factual questions as non-sensitive to reduce false-positive sensitive predictions. 

\noindent\textbf{Named Entity as an Atomic Phrase: }
Finally, we leverage the NER modules to extract named entities from the user utterances and treat them as atomic phrases during matching sensitive phrases. This approach is particularly useful for the cases such as ``James Bond'', where the token ``Bond'' as a part of the atomic named entity no longer matches the financially sensitive word ``bond''.

\noindent\textbf{Short Jokes: }
In conversations, joking is a good strategy to ease tensions. We also observe that some users voluntarily request jokes. In order to avoid being classified as a sensitive response, we manually collect 272 clean and harmless jokes for kids online \footnote{\url{https://redtri.com/best-jokes-for-kids/slide/33}} \footnote{\url{https://www.ducksters.com/jokes/}}. All these jokes are in question-answer pairs that do not require particular contexts. Since we are discouraged by the Alexa Prize organizers to build stand-alone joke modules, we only respond with jokes to users when we detect consecutive offensive utterances from users to ease the tension.

\subsubsection{Ranking Strategy}
%KB0722
Within the CoBot framework, we have several response generators for different tasks: EVI (response generator for factual question answering), NRG (trained for generic conversation), and sensitive response generator (response for predefined sensitive questions). Meanwhile, we also added some custom response generators for other topics, e.g., the news module, movie CKT module, and mini CKT modules. As a result, CASPR generates multiple responses for a user utterance, however, only one of them should be used as a reply. Thus, the response candidates need to be ranked so that the best one can be returned. We exploited two kinds of ranking strategies in our bot: a neural-network-based ranking strategy and a rule-based strategy. The rule-based ranking strategy in CASPR ranks the result based on the intent and topic, e.g., if a user asks for some news about football, the result from the news module would rank higher than sports CKT. So, for each intent and topic, we created a dictionary to store the prioritized list of response generators. In other words, CASPR chooses the response whose response generator has higher priority based on the intent and topic of the user utterance. During our testing, we also found that the rule-based system may not always perform well as the intent and topic classification is not 100\% accurate. For this reason, as a fallback mechanism, we rely on the neural-network based ranking strategy. A neural-network-based API, that is included in the CoBot framework, yields a set of metrics to indicate whether a response is good for a user's utterance. The 5 result components from the API are ``comprehensible'', ``interesting'', ``engaging'', ``erroneous'', and ``on topic''. CASPR employed a polynomial equation which includes all the components to rank the candidate responses. The coefficient of the variables in the polynomial equation has been decided based on experimentation by our team.
%------

\subsection{Response Generators}
Response generator is the module responsible to act upon the dialog manager's input and returns candidate bot response(s). This module is made up of multiple task specific response generators. One or many response generators are activated based on the topic and user's intent and return prospective responses. For example, we have used the CoBot's out-of-the-box response generator - EVI as a general question-answering tool that deals with the user's question intent. We have added/modified a number of response generators to provide an appropriate and engaging response to a user's utterance. These are described next.

\subsubsection{Extending Knowledge of CASPR using Amazon Kendra}
We used the natural language question-answering capability of Amazon Kendra \cite{kendra} service as a strategy to increase the knowledge of CASPR. Amazon Kendra provides indexing capabilities on a wide variety of document formats. This idea works reasonably well where there is not many structured data available, such as the data we had available to build Movie CKT and Book CKT. 
We chose Covid-19 as the primary topic to return responses from Kendra. As Covid-19 contains much medically sensitive jargon, we white-listed the words associated predominantly with Covid-19 to allow CASPR to respond. To index Covid-19 topic, we used the already available CORD-19 Service \cite{bhatia2020aws}. CORD-19 is an index built on top of Kendra and is periodically updated with medical journal articles about Covid-19. The availability of Cord-19 offloaded the cost associated with maintaining a Kendra index on AWS. Nonetheless, we maintained an index of our own to add Wikipedia articles about topics or persons prompted by CASPR users. We also performed an analysis of how the EVI bot and the NRG bots respond to Covid related questions in comparison to the Kendra bot. We show the following table with questions ranging from very basic questions to questions of increasing technicality. Questions lower in the table are more technical.

\par \noindent
\begin{tabular}{|p{5.5cm}|p{2cm}|p{2cm}|p{2cm}|}
 \hline
  Question & EVI & Cord-19 & NRG \\
  \hline
 Where did the virus originate? & Correct answer & No answer & No answer \\
 What are the symptoms for covid? & Correct answer & No answer & No answer \\
 Should I wear a mask?  & No answer & No answer & No answer \\
 What is the efficacy of Pfizer vaccine? & No answer & No answer & Unrelated answer \\
 What is the mrna vaccine & Correct answer? & Correct answer & No answer \\
 What is the cytokine storm for Covid? & No answer & Correct Answer & No answer \\ 
 \hline
\end{tabular}
\par
As the questions get technical, the Kendra bot provides correct answers. This is not surprising as the document indexed by CORD-19 are from medical journals. However, the EVI bot performs well on basic questions. Therefore, Kendra can be augmented with EVI to provided answers to more nuanced questions in a certain field/topic of interest. 
\par 
Finally, we also intended to fetch trending topics from Twitter and index their news items into Kendra. However, having the knowledge in Kendra and ensuring a proper conversation about the knowledge are different problems. Guiding a conversation around a new topic such as the ones emerging on Twitter is non-trivial. Further, the sensitivity around a new topic was uncertain. Addressing these problems is part of our future work in the context of conversational bots.

\subsubsection{News Module}

\noindent\textbf{News Scraping and Storage: }
CASPR news module uses Washington Post News API as the main resource of news. CASPR uses Python scripts to get news periodically, then save them by using Elastic Search. Both the scripts and Elastic Search node run on EC2.

\begin{itemize}
    \item \textit{\textbf{\underline{Get News}}}: CASPR news module gets 50 news daily by calling Washington Post news API. Since the news resource may contain news in other languages (e.g., Spainish, Italian, etc.), CASPR sets a filter that only allows English news to pass. CASPR may not able to pronounce special characters (e.g., \#, \~, etc.), these are also filtered out. 
    
    The news is classified into four categories: politics, sports, sci-tech, and business. CASPR uses a pre-trained content classification module to sort the news into each category. CASPR also generates two summaries (short, long) for each news, the short one has only 1 sentence while the long one has 3 sentences.
    
    \item \textit{\textbf{\underline{Saving News}}}: Each news will be saved by Elastic Search. Each news record has its id, content, short summary, long summary, category. News id is generated by the timestamp when it is recorded. The content is the JSON file provided by the news API, which contains the headline, body, publish time, resource, keywords, etc. The summaries and categories are generated by our pre-trained module.
\end{itemize}

\begin{algorithm}[t]
    \small
    \caption{Movie CKT response generator}\label{algorithm1}
    \hspace*{\algorithmicindent} \textbf{Input:} \textit{$CF$}: \text{change movie flag}\\ 
    \hspace*{\algorithmicindent} \textbf{Input:} \textit{$movie_{current}$}: \text{current movie}\\ 
    \hspace*{\algorithmicindent} \textbf{Input:} \textit{$question_{last}$}: \text{last question}\\ 
    \hspace*{\algorithmicindent} \textbf{Input:} \textit{$stack_{topic}$}: \text{topic question stack}\\ 
    \hspace*{\algorithmicindent} \textbf{Output:} \textit{response to user}
    \begin{algorithmic}[1]
        \Procedure{movieCKTRG}{}
        \State \(topics\) \(\gets\) \([ACTOR, DIRECTOR, PLOT, REVIEW, RATING, AWARD, TRIVIA]\)
        \State \textbf{if} \(movie_{current}\) is \(null\) \textbf{then}
        \State \hspace{4mm} \(movie_{current}\) \(\gets\) \(random(movies_{seed})\)
        \State \hspace{4mm} \(stack_{topic}\) \(\gets\) \(shuffle(topics)\)
        \State \textbf{end if}
        \State \textbf{if} \(CF\) is \(True\) \textbf{then}
        \State \hspace{4mm} \textbf{if} intent is \(YES\) \textbf{then}
        \State \hspace{4mm} \hspace{4mm} \(phrase_{answer}\) \(\gets\) \(answer(question_{last})\)
        \State \hspace{4mm} \hspace{4mm} \(response\) \(\gets\) \(answer(phrase_{answer})\)
        \State \hspace{4mm} \textbf{end if}
        \State \hspace{4mm} \(movie_{next}\), \(phrase_{bridge}\) \(\gets\) \(get\_next\_movie()\)
        \State \hspace{4mm} \(response\) \(\gets\) \(response\) + \(phrase_{bridge}\)
        \State \hspace{4mm} \(movie_{current}\) \(\gets\) \(movie_{next}\)
        \State \hspace{4mm} \(stack_{topic}\) \(\gets\) \(shuffle(topics)\)
        \State \textbf{else if} \(question_{last}\) is not \(null\)
        \State \hspace{4mm} \(phrase_{answer}\) \(\gets\) \(answer(question_{last})\)
        \State \hspace{4mm} \textbf{if} \(question_{last}\) is \(GENERIC\) or intent is \(NO\) \textbf{then}
        \State \hspace{4mm} \hspace{4mm} \textbf{if} \(random(types)\) is \(1\) \textbf{then}
        \State \hspace{4mm} \hspace{4mm} \hspace{4mm} \(question_{followup}\) \(\gets\) \(generic\_question()\)
        \State \hspace{4mm} \hspace{4mm} \textbf{else}
        \State \hspace{4mm} \hspace{4mm} \hspace{4mm} \(question_{followup}\) \(\gets\) \(stack_{topic}.pop()\)
        \State \hspace{4mm} \hspace{4mm} \textbf{end if}
        \State \hspace{4mm} \textbf{else}
        \State \hspace{4mm} \hspace{4mm} \textbf{if} \(random(types)\) is \(1\) or \(stack_{topic}\) is \(\emptyset\) \textbf{then}
        \State \hspace{4mm} \hspace{4mm} \hspace{4mm} \(question_{followup}\) \(\gets\) \(generic\_question()\)
        \State \hspace{4mm} \hspace{4mm} \textbf{else}
        \State \hspace{4mm} \hspace{4mm} \hspace{4mm} \(question_{followup}\) \(\gets\) \(stack_{topic}.pop()\)
        \State \hspace{4mm} \hspace{4mm} \textbf{end if}
        \State \hspace{4mm} \textbf{end if}
        \State \hspace{4mm} \(response\) \(\gets\) \(phrase_{answer}\) + \(question_{followup}\)
        \State \textbf{end if}
        \(question_{last}\) \(\gets\) \(question_{followup}\)
        \State \textbf{return} \(response\)
        \EndProcedure
    \end{algorithmic}
\end{algorithm}

\noindent\textbf{News Query Handling:} 
The news module returns results according to different queries. When there is no specific topic provided, such as “Tell me some news”, CASPR will return a piece of random news. While a query with a keyword, for instance, “Tell me some news about baseball”, CASPR needs to return a piece of news that is related to that keyword.

\begin{itemize}
    \item \textit{\textbf{\underline{Random News Query: }}}
    When a topic-less query comes, the upper layer of CASPR will pass the news query with an empty topic value to the news module. After receiving the query, the news module will pick a random news item from the candidates, uniformly chosen from the four categories. If the chosen news cannot pass the offensive filter, the choosing process repeats. Otherwise, the chosen news is returned.
    \item \textit{\textbf{\underline{Keyword News Query}}}
    When a query comes with a keyword, the news module will receive a topic value (keyword). Then CASPR searches the keyword in Elastic Search, which returns all matched records. The news module then randomly picks one record, and if it passes the offensive filter, it is returned, otherwise, another piece is randomly chosen.
\end{itemize}

\begin{algorithm}[t]
    \small
    \caption{Outer Conversation Loop}\label{algorithm2}
    \hspace*{\algorithmicindent} \textbf{Input:} \textit{$topic_{current}$}: \text{current}\\ 
    \hspace*{\algorithmicindent} \textbf{Input:} \textit{$TS$}: \text{topic stack}\\ 
    \hspace*{\algorithmicindent} \textbf{Input:} \textit{$C$}: \text{counter}\\ 
    \hspace*{\algorithmicindent} \textbf{Output:} \textit{response to user}
    \begin{algorithmic}[1]
        \Procedure{OuterLoop}{}
        \State \(topics\) \(\gets\) \([MOVIE, BOOK, MUSIC, SPORT, TECH, PETS, FAMILY]\)
        \State \textbf{if} \(topic_{current}\) is \(\emptyset\) or \(C = 0\)  \textbf{then}
        \State \hspace{4mm} \(TS\) \(\gets\) \(shuffle(topics)\)
        \State \hspace{4mm} \(topic_{current}\) \(\gets\) \(TS.pop()\)
        \State \hspace{4mm} \(C\) \(\gets\) \(N\) \Comment{N > 0, is the limit number of conversation in a topic}
        \State \textbf{end if}
        \State \textbf{if} \(topic_{utterance}\) is not predefined \textbf{then}
        \State \hspace{4mm} \textbf{return} \(bestOf(Kendra(), EVI(), MultiNRG())\) \Comment{return the best response according to ranking strategy}
        \State \textbf{else}
        \State \hspace{4mm} \(C\) \(\gets\) \(C - 1\)
        \State \hspace{4mm} \textbf{return} \(topicCKT(topic_{current})\) \Comment{call topic CKT}
        \State \textbf{end if}
        \EndProcedure
    \end{algorithmic}
\end{algorithm}

\subsubsection{Conversational Knowledge Toolkit (CKT)}

The conversational knowledge toolkit is CASPR's most important response generator that controls a conversation on a topic and returns relevant  outputs interactively. As discussed earlier (in section 3), we have classified these CKTs into two types based on their importance from the past conversations and the availability of the knowledge. Books, Movies, Music, etc. are considered as the main CKTs where CASPR can talk about a lot of the things and normally users also have diverse opinions and preferences on these topics. For the simple short topic discussion such as Pokemon, pet, travel, etc., CASPR has mini-CKTs. The design philosophy of the CKTs can be found in Section 3. Here, we give an example of the Movie CKT. Movie-CKT initiates by asking the user his/her favorite movie and then starts talking about the movie. Please note that if a user does not provide a movie name or if our customized NER is not able to track the movie name, then CASPR initiates the conversation with its favorite movie. The idea is to talk about different attributes of a movie such as an actor, director, awards, etc., and when the stack of the attributes becomes empty then CKT pivots to another movie by the same actor/director. CASPR uses IMDB database to get all this knowledge. Algorithm \ref{algorithm1} shows how the flow works, from initializing a movie name to getting a response. The algorithm is self-explanatory. Also, note that this a single flow of a conversation from the movie-CKT and we iteratively call the Movie-CKT unless the user wants to talk about another movie or the outer-loop decides to go to another topic. Similarly, we have built the CKTs for other topics such as Music and Books. 

The mini-CKTs are smaller version of these CKTs, where the topics are decided based on rule-based parsing of user utterances. Outer-loop decides when to call a mini-CKTs depending on the user utterance. The data flow is very similar to the movie-CKT algorithm with a simplified attribute selection strategy and with no pivot-based transitions. An example is given in section 3.1.

\subsubsection{Outer Conversation Loop}
In CASPR, the outer-loop controls all the custom CKTs and also it is capable of calling Kendra, EVI, or Multi-NRG as a fallback scenario. Based on our design philosophy, the outer-loop decides which response generator (RG) to call and selects the best response using the ranking strategy. The outer loop implements a high-level logic that randomly selects a topic from a set of topics and based on that invokes the respective RGs (most of the time those are CKTs). Also, it judges from the user responses whether a user is feeling bored, and if so, it changes the topic. Algorithm \ref{algorithm2} describes the process of choosing a random topic and continuing the conversation on that topic up to a fixed number of utterances until the user wants to change the topic or stops. If any specific RG response is null, then it selects the best response from the Kendra, EVI and, Multi-NRG responses. The idea is described earlier in Section 3.1.   

\subsubsection{Multi-NRG}
The NRG model is a GPT2-based model provided by CoBot framework, which aims to provide generic and natural responses to users based on the conversation history. The Policy Driven NRG is an improved version that uses dialog act tokens to generate responses at sentence level. The input to this model is a list of dialogue contexts separated by sentence. For most cases, it generates a good conversation without in-depth knowledge.  We use the PD-NRG model as the fallback plan for cases that the other response generators do not provide a good response. However, the PD-NRG module cannot generate exactly the same responses with specific input, sometimes the result is not ideal. Therefore, we employed strategy where we called multiple instances of PD-NRG in parallel to generate multiple candidate responses concurrently. Then we applied our ranking strategy to output the best result. We call this enhanced strategy Multi-NRG.

\subsection{Response Builder}
Response builder is the last module in the CoBot pipeline and it is responsible for packaging and delivering the bot utterance to the Alexa Skills Kit API. The default Response Builder passes the chosen response to ASK. We have used the default response builder provided with the CoBot framework.

\subsection{State Manager}
The state manager provides the interface between the CoBot framework and the back-end storage. The default CoBot’s state manager stores per-session and per-turn information about user utterances and CASPR's internal state to the DynamoDB. Also, we use the state manager to store the user-name details as well as the logging information such as RG's latency, false negative results, etc.

\begin{figure}[b]
\centering
\begin{minipage}{.5\textwidth}
  \centering
  \includegraphics[scale=0.5]{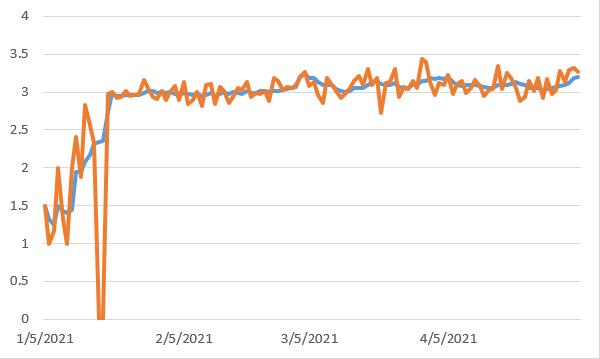}
      \caption{L1d \& L7d score}
    \label{fig:l1d}

\end{minipage}%
\begin{minipage}{.5\textwidth}
  \centering
    \includegraphics[scale=0.5]{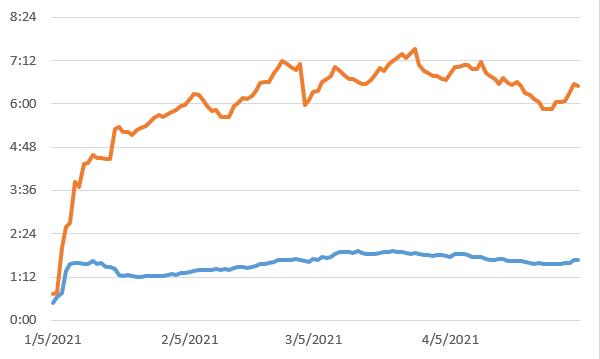}
    \caption{conversation duration}
    \label{fig:duration}
\end{minipage}
\end{figure}

\section{Results and Performance Evaluation}
Right at the beginning of the challenge when we started, we scored 1.5 for the L7d score (L7d is the average rating assigned by users over the last 7 days, L1d for last 1 day). After introducing more modules such as NRG for handling general utterances and EVI module for knowledge retrieval questions, our score gradually improved. 
%but around 17 Jan, the L1d score dropped to 0 because of a  critical error with the new added modules. After solving the critical issue, 
As we added the news module and the short-joke module to CASPR to handle more varied user utterances, our score improved further to around 3. The short-joke module especially helped when users became frustrated because CASPR would give a generic response if a user spoke a sensitive word. Telling a one-line joke makes the user less frustrated and to not stop the conversation immediately in such situations. 

% \noindent
%\setlength{\belowcaptionskip}{-10pt}
\begin{figure}[b!]
    \centering
    \includegraphics[width = \textwidth]{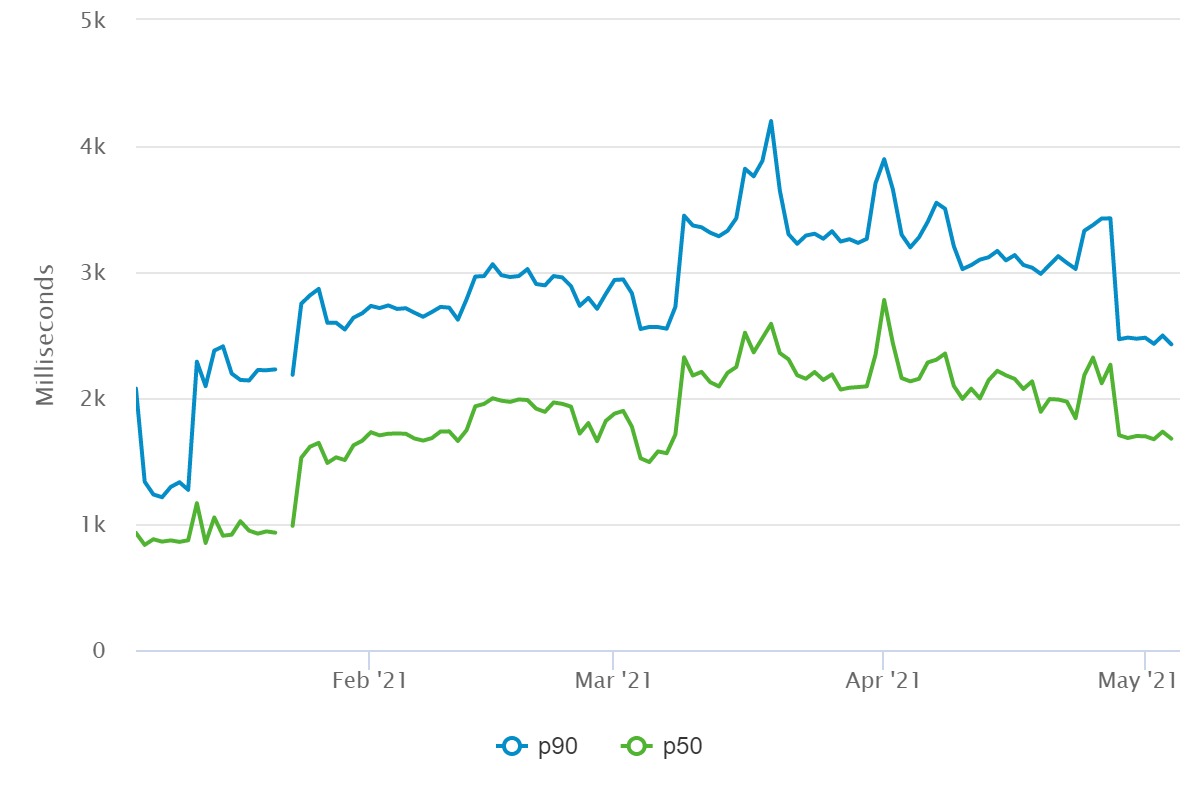}
    \caption{Response latency}
    \label{fig:latency}
\end{figure}

Figure \ref{fig:l1d} shows our L1d and L7d score with an orange curve and a blue curve, respectively. Figure \ref{fig:duration} shows  the average duration of user conversations on CASPR. Our 90th percentile came to around 5 minutes 30 seconds when the L7d reached around 3. CASPR's performance improved further as more modules (e.g., the movie CKT, mini-CKTs for various topics) were introduced to expand the depth and width of conversation. At the end of March, we added a new NRG-based response strategy and several neural-network-based modules for NER and sensitive word classification. 
These additions improved the conversation quality but lead to increases in latency as well as drop in conversation duration. We noticed this phenomenon, and removed most of the unnecessary logic, and replaced it with customized modules trained with smaller datasets, resulting in reduced latency. Adding the movie CKT greatly improved the quality of conversation related to movies and a corresponding increase in our score. Due to lack of time, we could not incorporate the book CKT and music CKT in CASPR and measure their impact on our L1d and L7d scores. Our belief is that these two CKTs would have led to an increase in L7d and L1d scores.

Figure \ref{fig:latency} shows the response latency throughout the challenge. Latency increased as new modules and functionality were added. There were several latency bumps: the first one happened in February because of the introduction of Spacy library \cite{spacy} for generic NER tasks. The second one happened at the end of March, caused by a new NRG strategy introduced in CASPR that generated more response candidates. The third one happened around April 20, 2021, as the movie CKT module added some latency due to the database access and complex API calling patterns. However, we could optimize these latencies and reduce them. Latency time had a direct correlation with  users' score---the greater the latency, the lower the score.

\medskip\noindent{\bf Testing during Development:}
For a specific module or functionality, one only has limited test cases with a narrow range. During the challenge period, we included a large number of beta testers 
%(CASPR team members, their friends, family members) 
to reduce our bias during testing. After each update with functionalities or new strategies, testers would test the newly added modules via mobile apps or devices using test-cases they have designed, then provide us detailed feedback, including corner cases encountered, their opinion about the updates, etc. Improvements were then made based on their feedback. For example, our testers told us that the news response became slower over time. This happened due to large number of news articles accumulating after two months of deployment. We fixed this by only considering news within the last 30-day window. After the deployment of the movie CKT, some testers gave feedback that the latency for some movie queries was intolerable or even timed out. Based on the provided context, we found the reason was partly that the database table was too large and partly the SQL queries we used were not efficient. Therefore, we reduced the number of movies in the database as well as optimized our SQL code. 

\medskip\noindent{\bf Analysis of Users' Conversation:}
We actively monitored the conversations of the users who were accessing CASPR. Every week we collected the conversations that happened during the last 7 days from the AWS database. We analyzed the reason why a user gave us a good/bad rating for a conversation. Bad ratings were a result of the natural language understanding problems, an unrelated response being given, or a conversation being repeated. Many users would say something offensive in this scenario, quit the conversation, and give us a rating of 1. To mitigate users' frustration we decided to employ one-liner jokes or serve random news pieces as transitional sentences and direct the users to other topics. This strategy helped in reducing the rate at which users were quitting the conversation. Thanks to these strategies, we could find what a user really valued, and then add new modules to react to users' interests. Some users also asked questions about COVID 19, but we were not able to give accurate answers to those questions with the EVI module, e.g., the latest COVID statistics. To answer questions related to COVID 19, we turned to Amazon's Kendra facility trained on COVID-related documents. The quality of the answers was decent but response latency was an issue.

\section{Accomplishments and Challenges}

Our CASPR team accomplished a lot in a short time, especially, given that it was one of the new teams, that the competition was being held during a pandemic, and our team was largely focused on using an approach based on commonsense reasoning for understanding utterances rather than the traditional machine learning-based approaches. Given that our approach departed so much from the traditional language models-based approaches that had to be integrated into the CoBot framework with various APIs, AWS, etc., involved, it took us longer to get most of the tasks accomplished. Despite these challenges, CASPR achieved good scores. We are quite hopeful that our continued research effort will produce a socialbot that can perform close to how a human would perform.

Some of the challenges we faced were the  software engineering task of working with the large CoBot framework, including learning the various APIs and the AWS infrastructure. Debugging was also hard due to the large software infrastructure involved. Keeping the latency down to a reasonable level was a major challenge as well, as users would get frustrated if the response was slow. Malicious users were also a challenge. They frequently accessed our bot then immediately exited, while giving a score of 1. Finally, integrating systems such as s(CASP) into the CoBot framework was quite challenging. In fact, integration issues, as well as latency issues prevented us from using s(CASP) for commonsense reasoning, and the s(CASP) system's capabilities had to be approximated in the various CKTs that we designed.

%declarative stuff; also automation in CKT
A major lesson learned is that representing knowledge in a declarative manner results in major advantages. We adopted the philosophy that code should be generated automatically as much as possible rather than be written. The design of CKTs and mini-CKTs greatly helped in this regard. Once the CKT for movies was designed, obtaining CKTs for books and music became quite an easy task. Similarly, for mini-CKTs, the knowledge for each topic was expressed declaratively, and then code quickly synthesized to implement that CKT.

\section{Future Work}

Developing conversational socialbots is a challenging project that employs almost most aspects of AI (speech recognition, speech generation, machine learning, natural language processing, automated reasoning, etc.). At present, most of the effort in the field is directed towards the use of machine learning, where language models are used to compute a response to a user utterance. These language models are very powerful: by using the language models provided in the CoBot framework and some of our own, plus a little bit of engineering, allowed us to reach a score of 3 quite effortlessly. However, as discussed earlier, ML-based response generation technology does not have any understanding of user's utterances. A correct response is more of a chance than a certainty. We strongly believe that for achieving a score of 4 and beyond, commonsense reasoning based techniques have to play a bigger role. 

Our goal is to develop conversational AI bots that ``understand'' user utterances, and then use the knowledge entailed in the utterance, along with its commonsense knowledge accumulated, to come up with a an effective response \cite{aaai21,basu2020square,discasp,aqua, basu_conv_iclp_dc}. We learned a lot during the development of the CASPR socialbot. Our understanding of how an end-to-end socialbot works significantly improved. Given our reliance on commonsense knowledge, which has to be created manually (tools such as ConceptNet \cite{conceptnet} and Microsoft Knowledge-graph \cite{mcg} are not based on modeling defaults, exceptions, and preferences), our goal is to develop domain-specific task bots first, e.g., simulate front-desk help, say, at our CS Department. For example, if a student walks in and wants to know his/her TA's office, but has no idea of the course's number or the instructor's name, the bot should be still able to correctly direct the student (for example, by holding a dialog and finding out what time the class is, and which room the class is held in, etc.).  

We are also developing new algorithms that are tailored for conversational AI for computing answer sets of an answer set program. We have developed the DiscASP algorithm \cite{discasp} for computing the set of responses that are relevant to a user utterance, with commonsense knowledge expressed in ASP. For example, if a user mentions a movie, then the DiscASP algorithm will compute that the socialbot can talk next about director, actors, trivia, movie-plot, awards, etc., of the movie. The DiscASP algorithm is a general algorithm that computes \textit{relevant consistent concepts} for a given topic, i.e., it finds those concepts that are closely related to the topic that the user is currently talking about. The socialbot can base its response on one or more elements of the computed relevant consistent concepts set.

We also plan to work on automating the acquisition of commonsense knowledge. Towards that end, we have been working on converting English text into ASP code \cite{basu2020square,aaai21}. The idea is to crowd-source commonsense knowledge in (a restricted subset of) English text and then convert it into defaults, exceptions, and preferences coded in ASP. 

Subsequently, we do plan to build an effective socialbot based on automated commonsense reasoning, ASP, and s(CASP) technology that exhibits close to human performance. A significant amount of commonsense knowledge about movies, books, music, and other conversational topics would have to be developed and added. However, we hope that our experience with the CASPR socialbot will help us in achieving our goal.

\section{Conclusion}

In this paper, we reported on the design and development of the CASPR socialbot, a conversational AI system to hold conversation with a human on general topics such as movies, books, music, sports, etc. Our goal was to use automated commonsense reasoning based on technologies such as answer set programming to ``understand'' users' utterances and to compute a response. Given the time constraints due to the competition and software engineering challenges due to having to deal with a large software infrastructure, we had to approximate the commonsense reasoning through design of techniques such as conversational knowledge template. CASPR was reasonably successful and achieved good scores. We plan to continue developing both task-specific bots as well as socialbots using the answer set programming and s(CASP) technology.

\subsubsection*{Acknowledgments}

The CASPR team would like to thank Amazon for giving it the opportunity to participate in the competition. The Amazon team's support and help were wonderful and indispensable. We especially would like to thank Kate Bland, Shui Hu, Anna Gottardi, Pankaj Rajan, Savanna, Yang Liu, Premraj Rengaram, Anjali Chadha, Anju, and Ali Binici for their help and guidance. The authors would also like to acknowledge support from Amazon and NSF grants IIS 1718945, IIS 1910131, IIP 1916206.

\bibliographystyle{plain}
\bibliography{main.bib}
\end{document}